\pdfoutput=1

\documentclass[11pt]{article}

\usepackage{acl}

\usepackage{times}
\usepackage{latexsym}
\usepackage{graphicx}
\usepackage{booktabs}
\usepackage{multicol}
\usepackage{multirow}
\usepackage{amsmath}
\usepackage{amssymb}
\usepackage{enumerate}
\usepackage{enumitem}
\usepackage{balance}

\usepackage[T1]{fontenc}

\usepackage[utf8]{inputenc}

\usepackage{microtype}

\title{KGAP: Knowledge Graph Augmented Political \\ Perspective Detection in News Media}

\author{Shangbin Feng$^\spadesuit$\thanks{\ \ These authors contributed equally to this work.} \: \: 
Zilong Chen$^{\spadesuit}$\footnotemark[1] \: \: 
Wenqian Zhang$^{\spadesuit}$\footnotemark[1] \: \: \\ \bf Qingyao Li$^{\spadesuit}$ \: \:
Qinghua Zheng$^{\spadesuit}$ \: \:
Xiaojun Chang$^{\diamondsuit}$ \: \:  Minnan Luo$^{\spadesuit}$\thanks{\ \ Corresponding author.}
\\
Xi’an Jiaotong University$^{\spadesuit}$ \: \: University of Technology Sydney$^{\diamondsuit}$ \\ \texttt{\{wind\_binteng,luoyangczl,2194510944,ly890306\}@stu.xjtu.edu.cn}\\ \texttt{qhzheng@mail.xjtu.edu.cn} \:  \texttt{cxj273@gmail.com} \:  \texttt{minnluo@xjtu.edu.cn}}

\begin{document}
\maketitle
\begin{abstract}
Identifying political perspectives in news media has become an important task due to the rapid growth of political commentary and the increasingly polarized political ideologies. Previous approaches focus on textual content and leave out the rich social and political context that is essential in the perspective detection process. To address this limitation, we propose KGAP, a political perspective detection method that incorporates external domain knowledge. Specifically, we construct a political knowledge graph to serve as domain-specific external knowledge. We then construct heterogeneous information networks to represent news documents, which jointly model news text and external knowledge. Finally, we adopt relational graph neural networks and conduct political perspective detection as graph-level classification. Extensive experiments demonstrate that our method consistently achieves the best performance on two real-world perspective detection benchmarks. Ablation studies further bear out the necessity of external knowledge and the effectiveness of our graph-based approach.
\end{abstract}

\section{Introduction}
The past decade has witnessed dramatic changes of political commentary~\citep{WILSON2020223} in two major ways. Firstly, the popularity of social media has led to a dramatic increase in political discussions on online media and social networks. Besides, ever-increasing political polarization has made it hard for journalists and news agencies to remain impartial in nature. Detecting political perspectives in the news media would help alleviate the issue of ``echo chamber'' \citep{barbera2015tweeting}, where only a single viewpoint is reiterated in closely formed communities and further deepens the divide. That being said, political perspective detection has become a pressing task which calls for further research efforts.

Previous news bias detection methods have focused on analyzing the textual content of news articles. Recurrent neural networks~\citep{HLSTM} and pre-trained language models~\citep{devlin2018bert} are adopted by ~\citet{li2021mean} to analyze news content for perspective analysis. ~\citet{CNNglove} uses convolutional neural networks with GloVe word embeddings \citep{pennington2014glove} for political perspective detection and achieves the best result in the hyperpartisan news detection task in SemEval 2019~\citep{SemEval}. \citet{li2021mean} leverages the attention mechanism and entity mentions in text and achieves state-of-the-art results.

\begin{figure}[t]
    \centering
    \includegraphics[width=\linewidth]{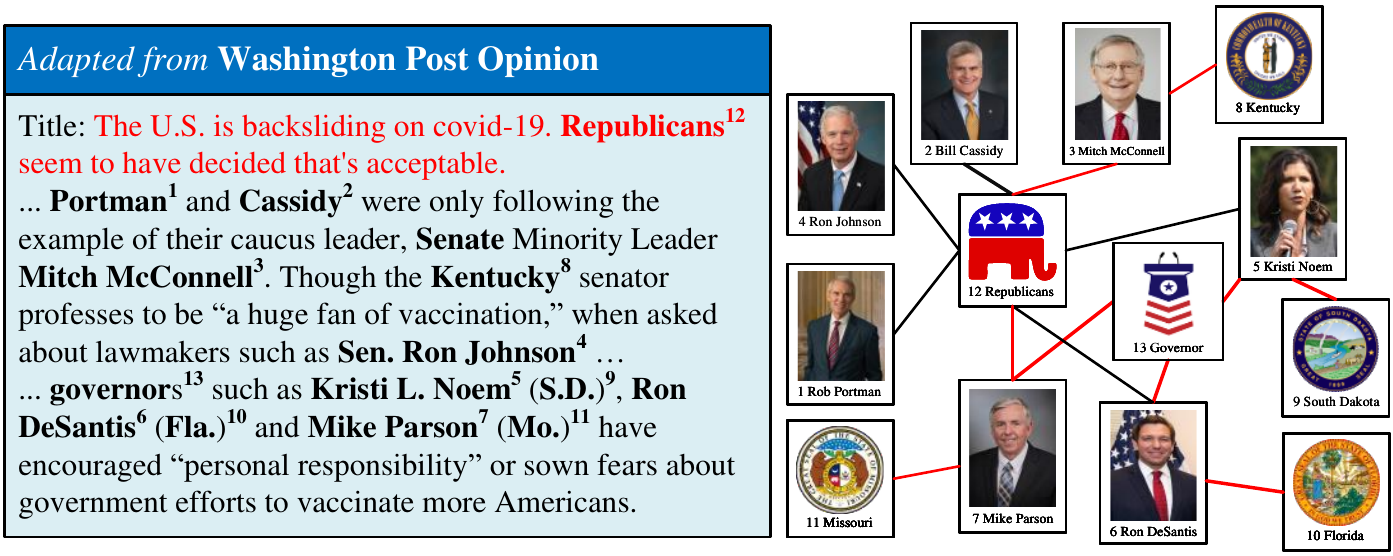}
    \caption{External knowledge of social and political entities in news articles that is essential for political perspective detection.}
    \label{fig:teaser}
\end{figure}

However, these text-based methods fail to incorporate pervasive social and political context in news articles, while human readers rely on external knowledge as background information to facilitate commonsense reasoning and perspective detection. For example, Figure \ref{fig:teaser} presents a news article that contains real-world entities such as political organizations, elected officials and geographical locations in the United States. External knowledge of these entities informs the reader that the mentioned individuals are republicans and mostly come from conservative states, which helps to identify the liberal stance expressed in criticizing them. This reasoning demonstrates that political perspective detection relies on social and political context as background information in addition to news content. 
That being said, political perspective detection methods should leverage external knowledge to reason beyond text and identify implicit perspectives.

\begin{figure*}[t]
    \centering
    \includegraphics[width=\linewidth]{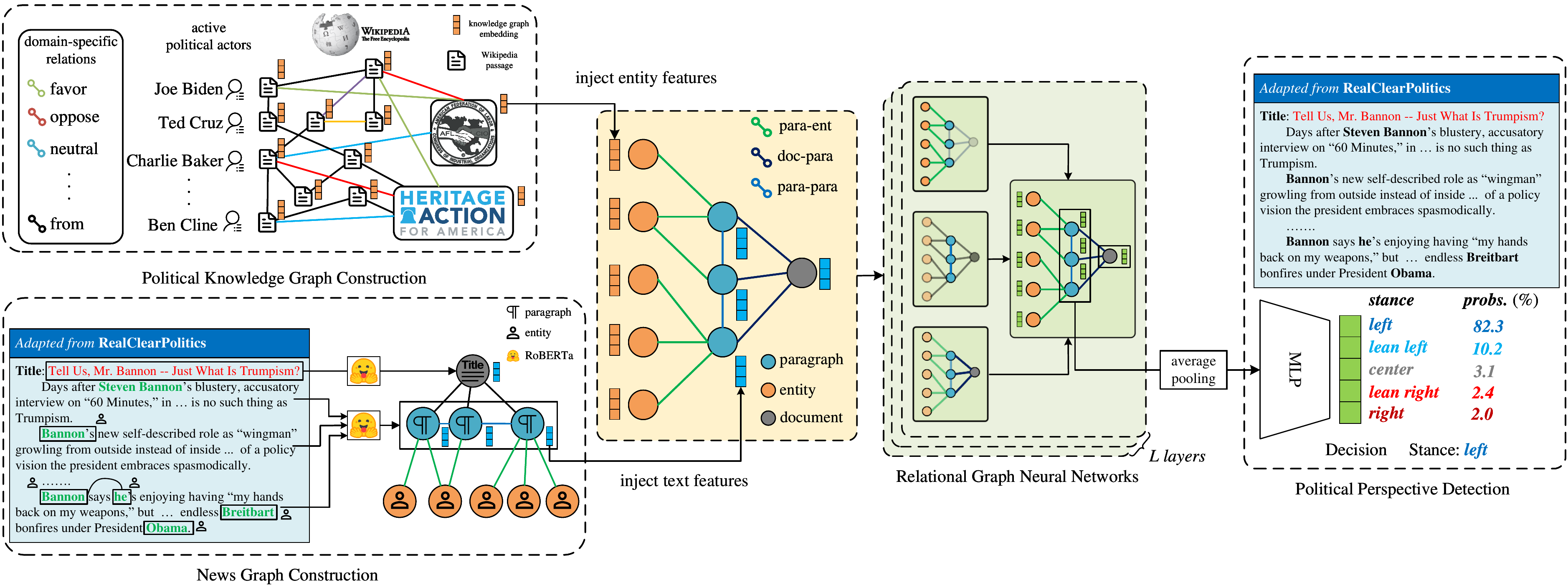}
    \caption{Overview of our proposed approach KGAP.}
    \label{fig:overview}
\end{figure*}
In light of these challenges, we propose \textbf{KGAP} (\textbf{K}nowledge \textbf{G}raph \textbf{A}ugmented \textbf{P}olitical perspective detection),  which leverages domain-specific external knowledge to incorporate background information and augment argument reasoning. Specifically, we firstly construct a political knowledge graph to represent the external knowledge that serves as background information for political narratives. We then construct heterogeneous information networks to represent news documents, which jointly model textual content and external knowledge in the knowledge graph. Finally, we adopt relational graph neural networks for graph representation learning and conduct political perspective detection as graph-level classification. Our main contributions are summarized as follows:
\begin{itemize}[leftmargin=*]
    \item To the best of our knowledge, we construct the first domain-specific knowledge graph of contemporary U.S. politics with 1,071 entities and 10,703 triples to serve as external knowledge for political perspective detection. We will publicize the political knowledge graph upon acceptance to facilitate research in related tasks.
    \item We propose to leverage domain knowledge and model news articles as heterogeneous information networks for political perspective detection. Our approach is end-to-end, inductive, and effectively framing the task as graph-level classification to incorporate external knowledge.
    \item We conduct extensive experiments to evaluate KGAP and competitive baselines. KGAP consistently outperforms state-of-the-art methods on two real-world datasets. Further studies demonstrate the necessity of external knowledge and explore the effect of knowledge graphs, text models, and the news graph structure in KGAP.
\end{itemize}

\section{Related Work}

\subsection{Political Perspective Detection}
The task of political perspective detection is generally studied in two different settings: social media and news media. For stance detection in social media, many works have studied the problem as text classification to identify stances in specific posts based on their textual content. They used techniques such as sentiment analysis~\citep{CIKMstance17} and neural attention networks~\citep{CIKMstance9}. Other approaches explored the problem of identifying stances of social media users instead of individual posts. ~\citet{CIKMstance7} conducted user clustering to identify their stances. ~\citet{CIKMstance23} proposed to predict user perspectives on major events based on network dynamics and user interactions.

For political perspective detection in news media, news documents are often classified into several perspective labels. Previous works have proposed to leverage semantic information in news articles with bias features~\citep{MEANbiasfeature} and various deep language encoders~\citep{li2021mean,HLSTM,CNNglove}. Later approaches have tried to enrich the textural content with the graph structures of online communities that interact with different news outlets. \citet{baly2020we} proposes to leverage media sources and design a triplet loss for political perspective detection. \citet{li2019encoding} supplemented news articles with a network of Twitter users and their interaction. In this paper, we focus on political perspective detection in news media and propose to incorporate political external knowledge.

\subsection{Knowledge-aware NLP}
Knowledge graphs (KGs) are often leveraged as external knowledge sources in various tasks in natural language processing. \citet{yu2021kgqa} leverages knowledge graphs for open-domain question answering.
\citet{hu2021knowledgefakenews} incorporates KGs in fake news detection with graph neural networks, knowledge graph embedding techniques and Wikipedia description of KG entities. \citet{liu2021knowledgestance} proposes to extract KG subgraphs and combine textual context and mentioned KG entities as input to language models. \citet{zhu2021dialogueemotiondetection} fuses commonsense statements in KGs with textual context using a transformer-based encoder-decoder architecture. In this paper, we construct a politics-specific knowledge graph and build on these works to incorporate political KGs into stance detection. Specifically, we explore the novel approach of leveraging knowledge graph embeddings \citep{TransE} as initial node features to enable knowledge-text interaction from a structural perspective.

\section{Methodology}
Figure \ref{fig:overview} presents an overview of our knowledge-aware political perspective detection approach KGAP. 
Specifically, we firstly construct a domain-specific knowledge graph of contemporary politics to serve as external knowledge for perspective detection. We then transform news articles into heterogeneous information networks, which jointly models textual content and related external knowledge. Finally, we adopt gated relational graph convolutional networks to learn graph representations and conduct political perspective detection.

\subsection{Knowledge Graph Construction}
Existing knowledge graphs (KGs) such as ConceptNet \citep{speer2017conceptnet}, Freebase \citep{bollacker2008freebase}, and YAGO \citep{tanon2020yago} have been leveraged in various NLP tasks such as misinformation detection \citep{hu2021knowledgefakenews}, knowledge-aware question answering \citep{yu2021kgqa}, and emotion detection \citep{zhu2021dialogueemotiondetection}. However, the task of political perspective detection demands politics-related external knowledge to facilitate argument reasoning, while existing KGs are too generic to be compatible or effective. In this paper, we propose to construct a political knowledge graph and hope to complement the scarce literature in domain-specific knowledge graph construction. The design goals of our political knowledge graph are summarized as follows:
\begin{itemize}[leftmargin=*]
    \item \textbf{domain-specific}: the knowledge graph should focus on closely related political background knowledge for efficient knowledge infusion.
    \item \textbf{diversified sources}: the knowledge graph should draw from both generic information sources and political expert knowledge.
    \item \textbf{scalable}: the knowledge graph construction process should be able to expand in scope to cope with unseen entities and task-specific settings.
    \item \textbf{adaptable}: the knowledge graph construction process should be applicable to different nations, time periods, and political systems.
\end{itemize}

Bearing these design principles in mind, we first select active U.S. political actors in the past decade as entities and determine ten types of political relations to ensure the KG is \textbf{domain-specific}. We then draw from Wikipedia pages of political actors as generic knowledge and two political think tanks, AFL-CIO\footnote{aflcio.org/scorecard} and Heritage Action\footnote{heritageaction.com/scorecard}, as political expert knowledge to determine triples and ensure the KG leverages \textbf{diversified sources}. Based on the initial entities and relations, we then leverage co-reference resolution \citep{lee2018higher} to expand the political KG by identifying new entities in Wikipedia documents. As a result, we construct a \textbf{scalable} KG while ensuring it's domain-specific for political perspective detection. Our political KG focuses on U.S. politics in the past decade to better align with news corpus, while our proposed construction process is \textbf{adaptable} to other nations and time periods by starting with corresponding political actors. As a result, we obtain a political knowledge graph with 1,071 entities and 10,703 triples to serve as external knowledge. A complete list of entities, relations and more KG construction details could be found in the technical appendix.

\subsection{News Graph Construction}
Graphs and graph neural networks have become increasingly involved in NLP tasks such as misinformation detection~\citep{hu2021knowledgefakenews} and question answering~\citep{yu2021kgqa}. In this paper, we construct news graphs to jointly model textual content and external knowledge. We also propose to leverage knowledge graph embeddings as node attributes to enable structural knowledge-text interaction. Specifically, we firstly determine the nodes in the news graph:
\begin{itemize}[leftmargin=*]
    \item \textbf{document node}: we use one document node to represent the news article. We use pre-trained RoBERTa \citep{liu2019roberta} to encode news title and use it as node attribute $v^d$.
    \item \textbf{paragraph node}: we use one node to represent each paragraph and use pre-trained RoBERTa for node features $v^p$.
    \item \textbf{entity node}: we use one node for each entity in our political KG. We propose the novel approach of using TransE \citep{TransE} KG embeddings as node attributes $v^e$.
\end{itemize}

After determining the nodes in the news graph, we construct a heterogeneous graph by connecting them with three types of edges:


\begin{itemize}[leftmargin=*]
    \item \textbf{doc-para edge}: the document node is connected with each paragraph node by doc-para edges. This ensures that every paragraph contributes to the overall perspective analysis.
    \item \textbf{para-para edge}: each paragraph node is connected with paragraphs that precede and follow it. These edges preserve the sequential flow of the original news document.
    \item \textbf{para-ent edge}: each paragraph node is connected with mentioned entities in our political KG. We conduct co-reference resolution and adopt TagMe \citep{tagme} to align text with entities. These edges aim to incorporate external knowledge for stance detection.
\end{itemize}

We denote these three edges as $R = \{doc-para, para-para, para-ent\}$. As a result, we obtain news graphs for each news article, which jointly model textual content and external knowledge to enable knowledge-text interaction from a structural perspective.
We then transform node attributes $v^d$, $v^p_i$, and $v^e_i$ with fully connected layers to obtain the initial node features, where $i$ denotes the $i$-th node in the set. Specifically, for document node and paragraph nodes:

\begin{equation}
    x_0^{(0)} = \phi(W_S \cdot v^d + b_S), \ \ \ x_i^{(0)} = \phi(W_S \cdot v^p_i + b_S)
\end{equation}
where $W_S$ and $b_S$ are learnable parameters and $\phi$ denotes Leaky-ReLU. Similarly, we use another fully connected layer to obtain initial features for entity nodes:
\begin{equation}
    x_i^{(0)} = \phi(W_E \cdot v^e_i + b_E)
                       \end{equation}
where $W_E$ and $b_E$ are learnable parameters. In the following, we use $x^{(0)}$ to denote initial node features for graph neural networks.

\begin{table*}[t]
    \setlength{\tabcolsep}{5mm}
    \centering
    \begin{tabular}{c c|c c|c c}
         \toprule[1.5pt] \multirow{2}{*}{\textbf{Method}} & \multirow{2}{*}{\textbf{Text+}} & \multicolumn{2}{c|}{\textbf{SemEval}} & \multicolumn{2}{c}{\textbf{AllSides}} \\ 
         & & \textbf{Acc} & \textbf{MaF} & \textbf{Acc} & \textbf{MaF} \\ \midrule[0.75pt]
         CNN\_GloVe & & $79.63$ & $N/A$ & $N/A$ & $N/A$ \\
         CNN\_ELMo & & $84.04$ & $N/A$  & $N/A$  & $N/A$  \\
         HLSTM\_GloVe & & $81.58$ & $N/A$  & $N/A$  & $N/A$  \\
         HLSTM\_ELMo & & $83.28$ & $N/A$  & $N/A$  & $N/A$  \\
         HLSTM\_Embed & & $81.71$ & $N/A$  & $76.45$ & $74.95$ \\
         HLSTM\_Output & & $81.25$ & $N/A$  & $76.66$ & $75.39$ \\
         BERT & & $84.03$ & $82.60$ & $81.55$ & $80.13$ \\
         MAN\_GloVe & \checkmark & $81.58$ & $79.29$ & $78.29$ & $76.96$ \\
         MAN\_ELMo & \checkmark & $84.66$ & $83.09$ & $81.41$ & $80.44$ \\
         MAN\_Ensemble & \checkmark & $86.21$ & $84.33$ & $85.00$ & $84.25$ \\
         \textbf{KGAP\_RGCN} & \checkmark & $89.22$ & $84.41$ & $\textbf{86.98}$ & $\textbf{86.53}$ \\
         \textbf{KGAP\_GRGCN} & \checkmark & $\textbf{89.56}$ & $\textbf{84.94}$ & $86.02$ & $85.52$ \\ \bottomrule[1.5pt]
    \end{tabular}
    \caption{Model performance on datasets SemEval and Allsides. $N/A$ denotes that the result is not reported in the related work. Text+ indicates whether the method leverages more than textual content. Acc and MaF denote accuracy and macro-averaged F1-score.}
    \label{tab:big}
\end{table*}

\subsection{Learning and Optimization}
We propose to adopt gated relational graph convolutional networks (gated R-GCNs) to learn representations for the news graphs and conduct political perspective detection as graph-level classification. 
For the $l$-th layer of gated R-GCN, we firstly aggregate messages from neighbors as follows:
\begin{equation}
    u_i^{(l)} = \Theta_{s} \cdot x_i^{(l-1)} + \sum_{r \in R} \sum_{j \in N_r(i)} \frac{1}{|N_r(i)|} \Theta_r \cdot x_j^{(l-1)}
\end{equation}
where $u_i^{(l)}$ is the hidden state for the $i$-th node in the $l$-th layer, $N_r(i)$ is node $i$'s neighborhood of relation $r$, $\Theta_s$ and $\Theta_r$ are learnable parameters. We then calculate gate levels:
\begin{equation}
    a_i^{(l)} = \sigma(W_A \cdot [u_i^{(l)}, x_i^{(l-1)}] + b_A)
\end{equation}
where $\sigma(\cdot)$ is the sigmoid function, $[\cdot,\cdot]$ denotes the concatenation operation, $W_A$ and $b_A$ are learnable parameters. We then apply the gate to $u_i^{(l)}$ and $x_i^{(l-1)}$:
\begin{equation}
    x_i^{(l)} = tanh(u_i^{(l)}) \odot a_i^{(l)} + x_i^{(l-1)} \odot (1 - a_i^{(l)})
\end{equation}
where $x_i^{(l)}$ is the output of the $l$-th gated R-GCN layer and $\odot$ denotes the Hadamard product operation. 

After applying a total of $L$ gated R-GCN layers, we obtain the learned node representations $x^{(L)}$. We then apply average pooling on paragraph node representations:
\begin{equation}
\label{equ:ours_AP}
    v^g = \frac{1}{s}\sum_{i=1}^{s} x_i^{(L)}
\end{equation}
where $s$ denotes the total amount of paragraphs in the news document. We then transform $v^g$ with a softmax layer:
\begin{equation}
    \hat{y} = softmax(W_O \cdot v^g + b_O)
\end{equation}
where $\hat{y}$ is the predicted perspective label, $W_O$ and $b_O$ are learnable parameters. We then derive the loss function $L$:
\begin{equation}
    L = -\sum_{D} \sum_{i=1}^Y y_i log(\hat{y_i}) + \lambda \sum_{w \in \theta} w^2
\end{equation}
where $Y$ is the number of stance labels, $y$ denotes stance annotation, $D$ represents the data set, $\theta$ denotes all learnable parameters in our proposed model, and $\lambda$ is a hyperparameter.

\section{Experiments}

\begin{table*}[]
    \centering
    \resizebox{1\linewidth}{!}{
    \begin{tabular}{c c c c c c}
         \toprule[1.5pt] \textbf{Dataset} & \textbf{\# News Articles} & \textbf{\# class} & \textbf{Class Distribution} & \textbf{Average \# Sentence} & \textbf{Average \# Word} \\ \midrule[0.75pt]
         \textbf{SemEval} & 645 & 2 & 407 / 238 & 27.11 & 494.29 \\
         \textbf{Allsides} & 10,385 & 3 & 4,164 / 3,931 / 2,290 & 49.96 & 1040.05 \\ \bottomrule[1.5pt]
    \end{tabular}
    }
    \caption{Details of two datasets SemEval and Allsides.}
    \label{tab:dataset_detail}
\end{table*}
\subsection{Dataset}
We evaluate our approach on the same benchmark datasets as in previous works \citep{li2019encoding, li2021mean}, namely SemEval and Allsides. Dataset details are are presented in Table \ref{tab:dataset_detail}. SemEval is the training data set from the SemEval 2019 Task 4: Hyperpartisan News Detection~\citep{SemEval}. Allsides is a larger and more diversified political perspective detection dataset collected in \citet{li2019encoding}. We follow the same settings in \citet{li2019encoding, li2021mean} so that our results are directly comparable with previous works. 

%

\subsection{Baselines}
We compare our method with the following competitive baselines on the SemEval and Allsides benchmarks:

\begin{itemize} [leftmargin=*]
    \item \textbf{CNN}~\citep{CNNglove} is the first place solution from the SemEval 2019 Task 4 contest~\citep{SemEval}. It uses GloVe (CNN\_GloVe) and ELMo (CNN\_ELMo) word embeddings with convolutional layers for stance prediction.
    \item \textbf{HLSTM}~\citep{li2019encoding} stands for hierarchical long short-term memory networks~\citep{HLSTM}. It encodes news with GloVe (HLSTM\_GloVe) and ELMo (HLSTM\_ELMo), hierarchical LSTMs, and aggregate with self-attention for political perspective detection.
    \item \textbf{HLSTM\_Embed} and \textbf{HLSTM\_Output}~\citep{li2021mean} use Wikipedia2Vec~\citep{yamada2018wikipedia2vec} and BERT-inspired masked entity models to learn entity representations and concatenate them with word embeddings or document representation for stance detection.
    \item \textbf{BERT}~\citep{devlin2018bert}  Pre-trained BERT is fine-tuned on the specific task of political perspective detection. 
    \item \textbf{MAN}~\citep{li2021mean} is a political perspective detection method that is pre-trained with social and linguistic information and fine-tuned with news bias labels.
\end{itemize}


\begin{figure}
    \centering
    \includegraphics[width=1\linewidth]{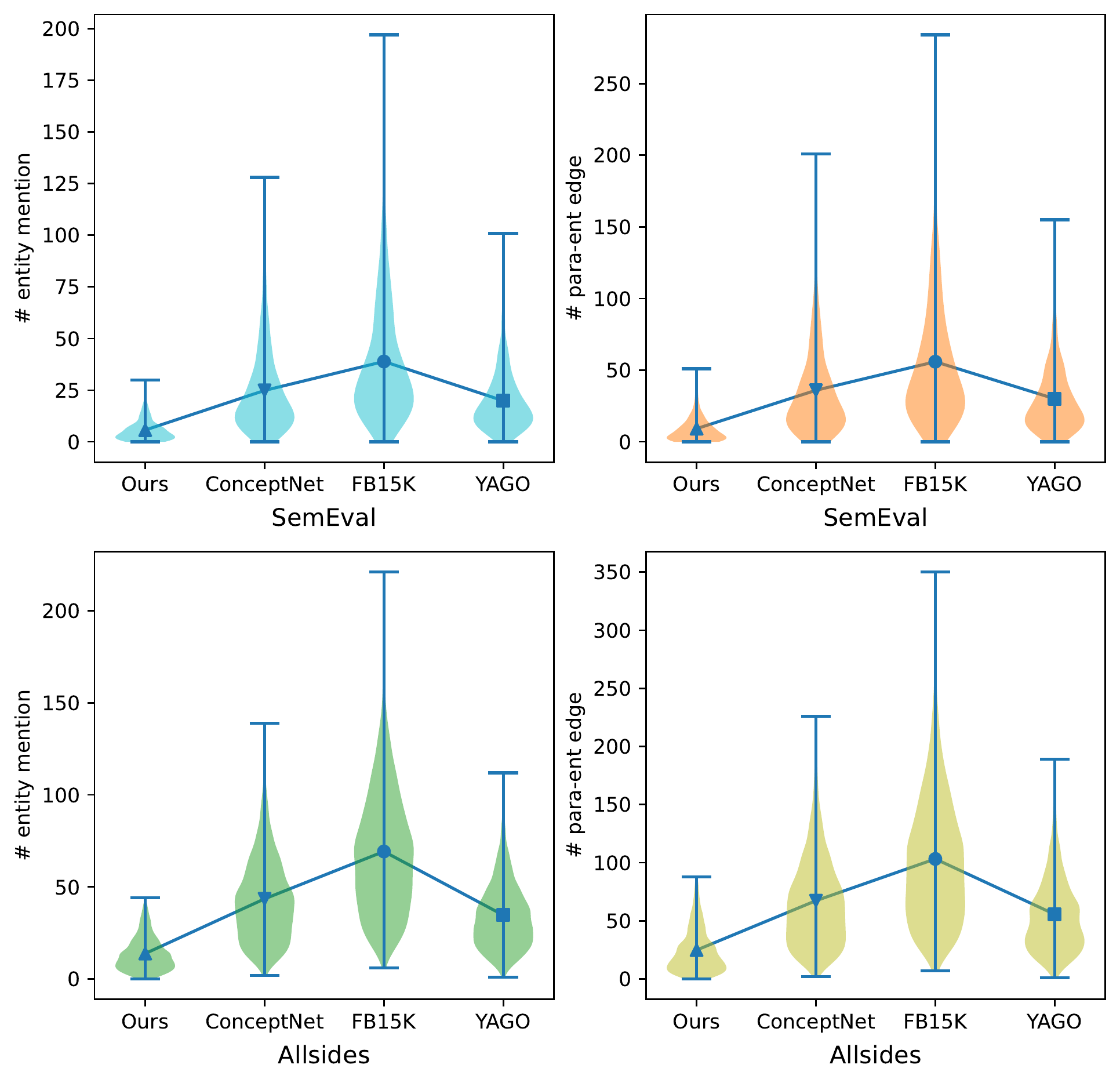}
    \caption{Statistics of entity mentions and para-ent edges under generic and domain-specific knowledge graphs. The line, shadow and line segment indicates the average, distribution and range of entity mentions and para-ent edges on two benchmark datasets.}
    \label{fig:knowldege_type_kg_stats}
\end{figure}

\subsection{Implementation}
We implement our proposed knowledge-aware political perspective detection approach with pytorch \citep{paszke2019pytorch}, pytorch lightning \citep{Falcon_PyTorch_Lightning_2019}, pytorch geometric \citep{torchgeometric}, and the HuggingFace transformers library \citep{wolf-etal-2020-transformers}. We present the hyperparameter settings of our proposed knowledge-aware political perspective detection approach in Table \ref{tab:hyperparameter} to facilitate reproduction. We follow the same hyperparameter settings in Table \ref{tab:hyperparameter} to compare with baselines and conduct ablation studies unless stated otherwise.

\begin{table}[]
    \centering
    \resizebox{0.9\linewidth}{!}{
    \begin{tabular}{c c}
         \toprule[1.5pt] \textbf{Hyperparameter} & \textbf{SemEval / Allsides} \\ \midrule[0.75pt]
         \ \ \ RoBERTa embedding size \ \ \  & \ \ \ 768 \ \ \ \\
         \ \ \ TransE embedding size \ \ \  & \ \ \ 200 / 768 \ \ \ \\
         \ \ \ gated R-GCN hidden size \ \ \  & \ \ \ 512 \ \ \  \\
         \ \ \ optimizer \ \ \  & \ \ \ Adam \ \ \  \\
         \ \ \ learning rate \ \ \  & \ \ \ $10^{-3}$ \ \ \  \\
         \ \ \ batch size \ \ \  & \ \ \ 16 \ \ \  \\
         \ \ \ dropout \ \ \  & \ \ \ 0.5 / 0.6 \ \ \  \\
         \ \ \ maximum epochs \ \ \  & \ \ \ 50 / 150 \ \ \  \\
         \ \ \ $L2$-regularization $\lambda$ \ \ \  & \ \ \ $10^{-5}$ \ \ \  \\
         \ \ \ gated R-GCN layer count $L$ \ \ \  & \ \ \ 2 \ \ \  \\ \bottomrule[1.5pt]
    \end{tabular}
    }
    \caption{Hyperparameter settings of our model.}
    \label{tab:hyperparameter}
\end{table}

\subsection{Experiment Results}
\label{subsec:bigexp}
We run KGAP with R-GCN (KGAP\_RGCN) and gated R-GCN (KGAP\_GRGCN) and report model performance in Table \ref{tab:big}, which demonstrates that:
\begin{itemize}[leftmargin=*]
    \item KGAP consistently outperforms all competitive baselines on both SemEval and Allsides, including the state-of-the-art method MAN~\citep{li2021mean}.
    \item Two methods that leverage information other than textual content, namely MAN and ours, outperform other baselines. Our method further incorporates external knowledge in political knowledge graphs and outperforms MAN. 
    \item For methods that use word embeddings, ELMo often outperforms their GloVe counterparts. This suggests that text modeling is still essential in political perspective detection.
\end{itemize}

\begin{figure*}[t]
    \centering
    \includegraphics[width=1\linewidth]{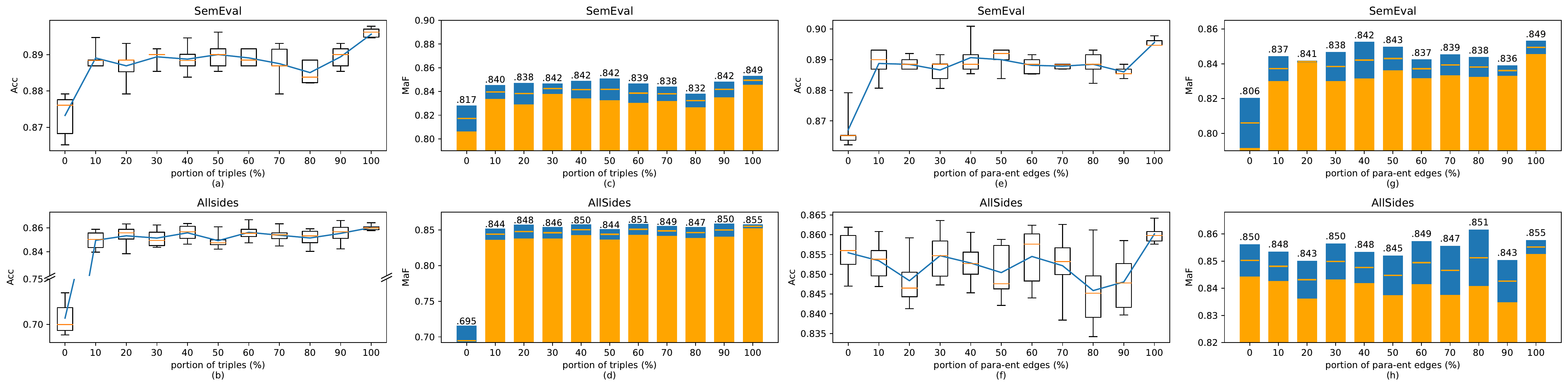}
    \caption{Average model performance and standard deviation when triples in our constructed political knowledge graph (a-d) and para-ent edges in our news graphs (e-h) are gradually removed.}
    \label{fig:knowledge_density_triple}
\end{figure*}

In conclusion, KGAP achieves state-of-the-art performance on two widely adopted benchmarks. In the following, we study the effect of external knowledge, textual content, and graphs in our method and the task of political perspective detection. We run all experiments \textbf{five times} to ensure a consistent evaluation and report the average performance as well as standard deviation if necessary.

\subsection{Knowledge Study}

\subsubsection{Knowledge Density}
We study the effect of knowledge density with two different settings. We gradually remove triples in our constructed political KG and report model performance in Figure \ref{fig:knowledge_density_triple} (a) - (d). We then gradually remove para-ent edges in news graphs and report model performance in Figure \ref{fig:knowledge_density_triple} (e) - (h). In both settings, model performance drops significantly with very little knowledge, while quickly saturating with $10\%$ to $20\%$ external knowledge. Another performance boost is often witnessed at $100\%$ external knowledge, which argues for complete KGs. Besides, the effect of external knowledge is greater on the smaller dataset SemEval, which suggests that our knowledge-aware approach is especially effective with limited data.

\subsubsection{Knowledge Type}
To examine whether our constructed political knowledge graph is essential in our model's performance, we compare it with various generic knowledge graphs \citep{speer2017conceptnet,bollacker2008freebase,tanon2020yago} that are widely adopted in NLP tasks. Using TagMe \citep{tagme} as a tool for text-entity alignment, Figure \ref{fig:knowldege_type_kg_stats} illustrates how many KG entities are mentioned in news articles and how many para-ent edges are captured in constructed news graphs. It shows that fewer entities and para-ent edges are provided by our politics-specific knowledge graph, which indicates that our KG adds little computational burden to the news graphs. We then report model performance paired with different knowledge graphs in Table \ref{tab:knowledge_type_kgs_performance}. It is demonstrated that our domain-specific knowledge graph outperforms generic knowledge graphs in political perspective detection, which indicates successful domain-specific KG construction and proves the necessity and efficiency of our political knowledge graph.


\begin{table}[t]
    \centering
    \begin{tabular}{c|c c|c c}
         \toprule[1.5pt] \multirow{2}{*}{\textbf{KG}} & \multicolumn{2}{c|}{\textbf{SemEval}} & \multicolumn{2}{c}{\textbf{AllSides}} \\ 
         & \textbf{Acc} & \textbf{MaF} & \textbf{Acc} & \textbf{MaF} \\ \midrule[0.75pt]
         ConceptNet & $87.81$ & $82.65$ & $85.59$ & $85.08$ \\
         FreeBase & $87.98$ & $82.73$ & $85.84$ & $85.32$ \\
         YAGO & $87.30$ & $81.78$ & $85.18$ & $84.66$ \\
         \textbf{Ours} & $\textbf{89.56}$ & $\textbf{84.94}$ & $\textbf{86.02}$ & $\textbf{85.52}$ \\ \bottomrule[1.5pt]
    \end{tabular}
    \caption{Average model performance when using generic and domain-specific knowledge graphs as external knowledge.}
    \label{tab:knowledge_type_kgs_performance}
\end{table}

\begin{table}[t]
    \centering
    \begin{tabular}{c|c c|c c}
         \toprule[1.5pt] \multirow{2}{*}{\textbf{KGE Method}} & \multicolumn{2}{c|}{\textbf{SemEval}} & \multicolumn{2}{c}{\textbf{AllSides}} \\ 
         & \textbf{Acc} & \textbf{MaF} & \textbf{Acc} & \textbf{MaF} \\ \midrule[0.75pt]
         TransE & $89.56$ & $84.94$ & $86.02$ & $85.52$ \\ 
         TransR & $88.54$ & $83.45$ & $85.15$ & $84.61$ \\
         DistMult & $88.51$ & $83.63$ & $84.47$ & $83.90$ \\
         HolE & $88.85$ & $83.68$ & $84.78$ & $84.24$ \\
         RotatE & $88.84$ & $84.04$ & $85.61$ & $85.11$ \\ 
         \bottomrule[1.5pt]
    \end{tabular}
    \caption{Average model performance with entity node attributes derived by different knowledge graph embedding techniques.}
    \label{tab:kge_methods}
\end{table}


\subsubsection{Knowledge Graph Embedding}
\begin{figure}[t]
    \centering
    \includegraphics[width=1\linewidth]{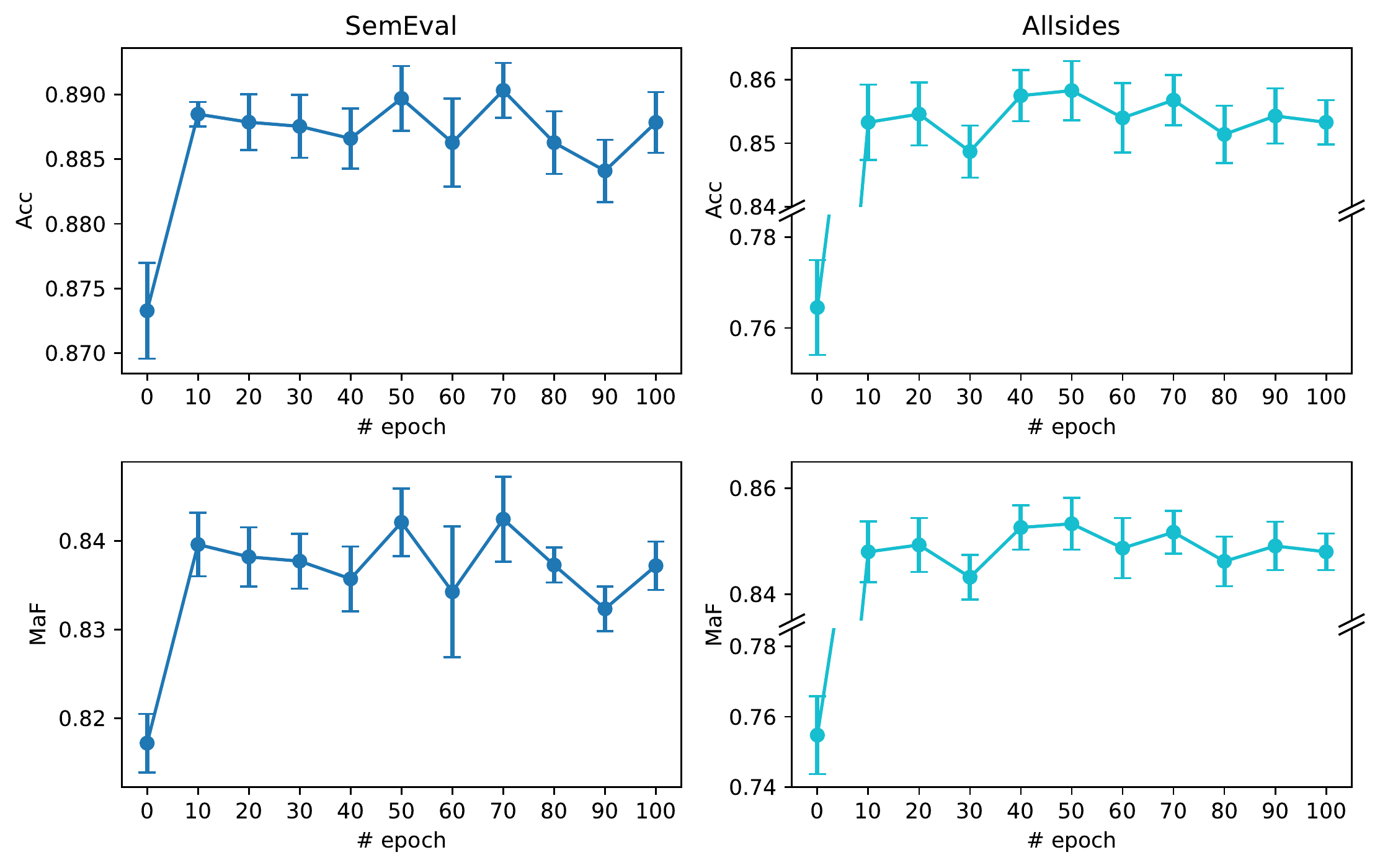}
    \caption{Average model performance and standard deviation when TransE KG embeddings are trained for different epochs.}
    \label{fig:kge_epochs}
\end{figure}

In this paper, we propose the novel approach of leveraging knowledge graph embeddings as initial node features to facilitate knowledge-text interaction from a structural perspective. We further explore this idea with other knowledge graph embedding techniques \citep{transr, distmult, hole, sun2019rotate} and report model performance in Table \ref{tab:kge_methods}. It is demonstrated that model performance on both datasets does not change significantly, indicating that our approach does not rely on specific knowledge graph embedding techniques. We further train TransE~\citep{TransE} to different extents and report performance in Figure \ref{fig:kge_epochs}. It is illustrated that as little as 10 epochs of TransE training would lead to better task performance than random initialization (0 epoch), while our approach is generally effective with knowledge graph embeddings with more training epochs.


\begin{figure}[t]
    \centering
    \includegraphics[width=1\linewidth]{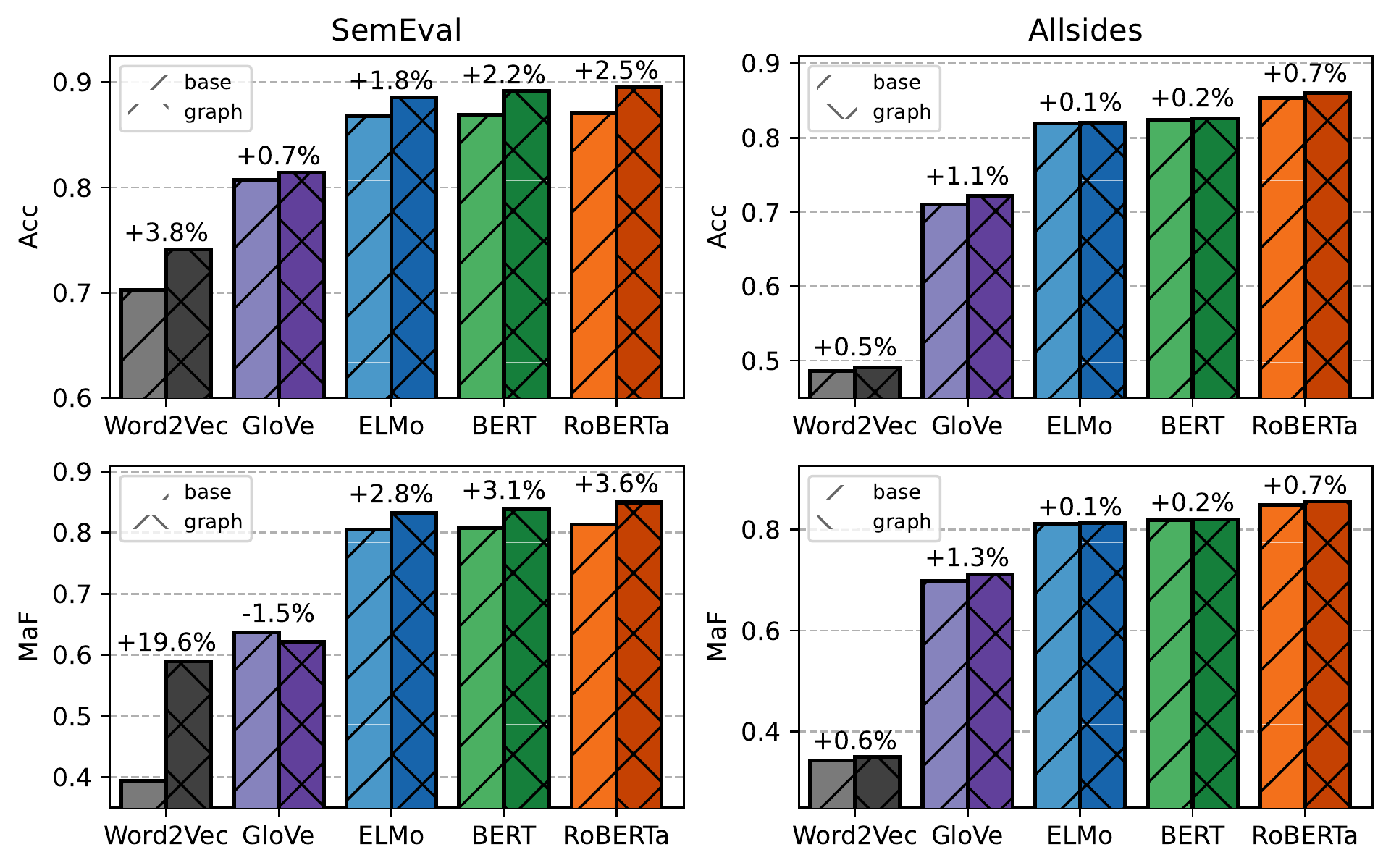}
    \caption{Average performance of text models (base, lighter) and combining them with our news graph structure (graph, darker).}
    \label{fig:text_analysis}
\end{figure}

\subsection{Text Study}
We observed in Section \ref{subsec:bigexp} that text modeling plays an important role in political perspective detection. We further explore how different text analysis techniques perform on their own and when combined with our news graphs. Specifically, we encode news text with text analysis techniques and conduct political perspective detection with fully connected layers for text-only settings. We then substitute the paragraph node attributes with these methods for graph-based settings. Figure \ref{fig:text_analysis} demonstrates that advanced language models consistently outperform classic methods, indicating the correlation between model performance and text processing ability. Besides, Figure \ref{fig:text_analysis} shows that language modeling methods perform better when combined with our proposed news graph, indicating the necessity of incorporating structural information and external knowledge in political perspective detection.

\subsection{Graph Study}
We propose the novel approach of constructing news graphs to jointly model textual content and external knowledge while conducting political perspective detection as graph-level classification. To examine the effect of news graphs, we firstly remove three types of edges and report model performance in Table \ref{tab:graph_edge}. It is demonstrated that all three types of edges contribute to model performance, while para-ent edges are most essential on the smaller SemEval since they help incorporate external knowledge to boost task performance on limited data. Besides, we adopt R-GCN and gated R-GCN for graph representation learning on our constructed news graphs. We further explore the effect of GNNs by changing to other GNN layers and report performance in Table \ref{tab:graph_gnn}, which indicates the necessity of GNN heterogeneity in our approach.

\begin{table}[t]
    \centering
    \resizebox{\linewidth}{!}{
    \begin{tabular}{c|c c|c c}
         \toprule[1.5pt] \multirow{2}{*}{\textbf{Ablation Setting}} & \multicolumn{2}{c|}{\textbf{SemEval}} & \multicolumn{2}{c}{\textbf{AllSides}} \\ 
         & \textbf{Acc} & \textbf{MaF} & \textbf{Acc} & \textbf{MaF} \\ \midrule[0.75pt]
         no doc-para edges & $89.03$ & $84.23$ & $-$ & $-$ \\
         no para-para edges & $88.41$ & $83.14$ & $82.52$ & $81.83$ \\
         no para-ent edges & $86.71$ & $80.60$ & $85.54$ & $85.02$ \\ 
         \textbf{full graph} & $\textbf{89.56}$ & $\textbf{84.94}$ & $\textbf{86.02}$ & $\textbf{85.52}$ \\ \bottomrule[1.5pt]
    \end{tabular}
    }
    \caption{Average model performance when three different types of edges in the news graphs are respectively removed. }
    \label{tab:graph_edge}
\end{table}

\begin{table}[t]
    \centering
    \resizebox{\linewidth}{!}{
    \begin{tabular}{c c|c c|c c}
         \toprule[1.5pt] \multirow{2}{*}{\textbf{GNN Type}} & \multirow{2}{*}{\textbf{Het.}} & \multicolumn{2}{c|}{\textbf{SemEval}} & \multicolumn{2}{c}{\textbf{AllSides}} \\ 
         & & \textbf{Acc} & \textbf{MaF} & \textbf{Acc} & \textbf{MaF} \\ \midrule[0.75pt]
         GCN & & $87.20$ & $81.63$ & $86.91$ & $86.43$ \\
         GAT & & $89.00$ & $84.41$ & $84.12$ & $83.33$ \\
         GraphSAGE & & $89.12$ & $84.47$ & $86.77$ & $86.24$ \\
         R-GCN & \checkmark & $89.22$ & $84.41$ & $\textbf{86.98}$ & $\textbf{86.53}$ \\
         Gated R-GCN & \checkmark & $\textbf{89.56}$ & $\textbf{84.94}$ & $86.02$ & $85.52$ \\ \bottomrule[1.5pt]
    \end{tabular}
    }
    \caption{Average model performance with various heterogeneous and homogeneous GNNs. Het. indicates heterogeneous GNNs.}
    \label{tab:graph_gnn}
\end{table}

\section{Conclusion}
In this paper, we propose an end-to-end, inductive and graph-based approach to leverage domain-specific external knowledge in political perspective detection. We firstly construct a domain-specific knowledge graph as external knowledge for political perspective detection. We then construct news graphs to jointly model news content and external knowledge. Finally, we adopt relational graph neural networks to learn graph representations and framing the task as graph-level classification. Extensive experiments demonstrate that KGAP consistently outperforms state-of-the-art methods on two widely adopted perspective detection benchmarks. Further ablation studies indicate the necessity of external knowledge and examine the effect of knowledge, text, and graph in KGAP.

\clearpage

\bibliography{custom}

\clearpage

\appendix

\section{Limitation}
We identified two minor limitations of KGAP:
\begin{itemize}[leftmargin=*]
    \item We noticed that there are only a few para-ent edges in the news graphs of certain news articles. We examined these news articles and found that some entity mentions are not captured by the entity linking tool TagMe. This issue might be addressed by using better entity linking tools.
    \item Political facts change over time, for example, politicians in the United States may represent different districts due to redistricting. Our knowledge graph does not model such changes for now, while this issue may be addressed by introducing temporal knowledge graphs.
\end{itemize}

\section{Knowledge Graph Details}

We construct a political knowledge graph (KG) of contemporary U.S. politics to serve as domain-specific external knowledge for political perspective detection, while our approach could be extended to political scenarios in other countries. Table \ref{tab:KGlist} presents entity and relation types in our KG.

For \textbf{elected office}, we consider the White House, House of Representatives, Senate, Supreme Court and governorship. For \textbf{time period}, we consider the 114th to 117th congress while our KG could be similarly extended to other time periods. We then retrieve the \textbf{presidents}, \textbf{supreme court justices}, \textbf{senators}, \textbf{congresspersons}, and \textbf{governors} that overlap with these time periods from Wikipedia\footnote{\url{https://www.wikipedia.org/}} and U.S. Congress website\footnote{\url{https://www.congress.gov/}}. For \textbf{states}, we consider the 50 U.S. states. For \textbf{political parties}, we consider the Democratic Party, the Republican Party and independents. For the first five types of relations (affiliated\_to, from, appoint, overlap\_with and member\_of), we conduct co-reference resolution to identify new entity mentions, connect the entities and expand the KG. For \textbf{political ideology}, we use two entities to represent liberal conservative values. We then resort to the legislator scoreboards at AFL-CIO\footnote{aflcio.org/scorecard} and Heritage Action\footnote{heritageaction.com/scorecard}. These scoreboards score U.S. legislators from 0 to 100 to evaluate how liberal or conservative they are. We partition the score range into strongly favor (90 - 100), favor (75 - 90), neutral (25 - 75), oppose (10 - 25) and strongly oppose (0 - 10) to connect entities with these five relations. Our political KG draws from both generic information sources and political expert knowledge.

As a result, we construct a political knowledge graph with 1,071 entities and 10,703 triples to serve as external knowledge, which is domain-specific, scalable, adaptable and draws from diversified information sources. Our political KG could also be helpful for related tasks such as disinformation detection. We submit the constructed knowledge graph as supplementary material to facilitate reproduction.

\begin{table*}[]
    \centering
    \resizebox{1\linewidth}{!}{
    \begin{tabular}{c|c|c|c}
         \toprule[1.5pt] \textbf{Entity Type} & \textbf{Example} & \textbf{Relation Type} & \textbf{Example} \\ \midrule[0.75pt]
         elected office & the U.S. Senate & affiliated\_to & (Joe Biden, affiliated\_to, Democratic Party) \\
         time period & 117th congress & from & (Ted Cruz, from, Texas) \\
         president & Joe Biden & appoint & (Donald Trump, appoint, Amy Coney Barrett) \\
         supreme court justice & Amy Coney Barrett & overlap\_with & (Joe Biden, overlap\_with, 117th congress) \\
         senator & Elizabeth Warren & member\_of & (Dianne Feinstein, member\_of, the U.S. Senate) \\
         congressperson & Nancy Pelosi & strongly\_favor & (Bernie Sanders, strongly\_favor, liberal\_values) \\
         governor & Ron DeSantis & favor & (Joe Cunningham, favor, liberal\_values) \\
         state & Massachusetts & neutral & (Henry Cuellar, neutral, liberal\_values) \\
         political party & Republican Party & oppose & (Lamar Alexander, oppose, liberal\_values) \\
         political ideology & liberal values & strongly\_oppose & (Ted Cruz, strongly\_oppose, liberal\_values) \\ \bottomrule[1.5pt]
    \end{tabular}
    }
    \caption{List of entities and relations in our collected political knowledge graph.}
    \label{tab:KGlist}
\end{table*}

\section{Dataset Details}
We evaluate our knowledge-aware political perspective detection approach on SemEval \citep{SemEval} and Allsides \citep{li2019encoding}, two widely adopted benchmarks in previous works. We present dataset details in Table \ref{tab:dataset_detail}. For SemEval, we follow the same 10-fold evaluation settings and the exact folds as in \citet{li2021mean}. For Allsides, we follow the same 3-fold evaluation settings and the exact folds as in \citet{li2021mean}, while a few news urls have expired and we could not retrieve the original news content, leading to minor differences. In this way, our method's performance are directly comparable with previous state-of-the-art approaches \citep{li2019encoding,li2021mean}.




\section{Experiment Details}

\subsection{GNN Setting}
We train our approach with both relational graph convolutional networks (R-GCN) and gated relational graph convolutional networks (gated R-GCN), while our approach does not rely on any specific GNN architecture. All experiments in Section 4.4, 4.5 and 4.6 are conducted with gated R-GCN.
\subsection{Figure 4}
For better visualization effects in Figure 4, we leave out news articles whose number of entity mentions or para-ent edges is five times greater than the average value. This approximation does not compromise the main conclusion that our domain-specific knowledge graph involves less computational burden while performing better than generic knowledge graphs.

\subsection{Table 4}
The Allsides benchmark does not explicitly provide news titles. As a result, there are no doc-para edges in the constructed news graphs for this dataset, and thus the ``no doc-para edges" setting could not be conducted on Allsides.

\subsection{Figure 6}
For Word2Vec word embeddings, we use the gensim library \citep{gensim}.
For GloVe word embeddings, we use the pre-trained vectors\footnote{https://nlp.stanford.edu/projects/glove/} from Stanford NLP.
For ELMo word embeddings, we use the AllenNLP implementation \citep{allennlpelmo}.
For BERT and RoBERTa, we use pre-trained models in the transformers library \citep{wolf-etal-2020-transformers} to encode news content and serve as textual features. For all the base settings, we use Leaky-ReLU and two fully connected layers with 512 as hidden layer size.

\end{document}